\documentclass{article}

\usepackage{times}

\usepackage{hyperref}

\usepackage[accepted]{icml2015}

\usepackage{url}

\usepackage{amsfonts,amstext,amsmath,amssymb}

\usepackage{longtable}

\usepackage{graphicx}

\usepackage{booktabs} 
\usepackage{multirow}

\icmltitlerunning{``The Sum of Its Parts'': Joint Learning of Word and Phrase Representations with Autoencoders}

\begin{document}

\twocolumn[
\icmltitle{``The Sum of Its Parts'':\\Joint Learning of Word and Phrase Representations with Autoencoders}

% It is OKAY to include author information, even for blind
% submissions: the style file will automatically remove it for you
% unless you've provided the [accepted] option to the icml2015
% package.
\icmlauthor{
R\'emi Lebret}{remi@lebret.ch}
\icmladdress{Idiap Research Institute, Martigny, Switzerland\\
\'Ecole Polytechnique F\'ed\'erale de Lausanne (EPFL), Lausanne, Switzerland}
\icmlauthor{Ronan Collobert$^{1}$}{ronan@collobert.com}
\icmladdress{Facebook AI Research, Menlo Park, CA, USA\\Idiap Research Institute, Martigny, Switzerland}
% You may provide any keywords that you 
% find helpful for describing your paper; these are used to populate 
% the "keywords" metadata in the PDF but will not be shown in the document
\icmlkeywords{Learning, Natural Language Processing, Autoencoder, Word Representation}

\vskip 0.3in % set to 0.6 instead of 0.3
]

\begin{abstract}
Recently, there has been a lot of effort to represent words in continuous vector spaces. 
Those representations have been shown to capture both semantic and syntactic information about words.
However, distributed representations of phrases remain a challenge.
We introduce a novel model that jointly learns word vector representations and their summation. 
Word representations are learnt using the word co-occurrence statistical information.
To embed sequences of words (i.e.\ phrases) with different sizes into a common semantic space, we propose to average word vector representations.
In contrast with previous methods which reported \emph{a posteriori} some compositionality aspects by simple summation, we simultaneously train words to sum, while keeping the maximum information from the original vectors.
%This allows sequences of words (i.e. phrases) with different sizes to be embedded into a common semantic space.
We evaluate the quality of the word representations on several classical word evaluation tasks, and we introduce a novel task to evaluate the quality of the phrase representations.
While our distributed representations compete with other methods of learning word representations on word evaluations, we show that they give better performance on the phrase evaluation.
Such representations of phrases could be interesting for many tasks in natural language processing.

\end{abstract}

\section{Introduction}

Human language ``makes infinite use of finite means"~\cite{humboldt1836verschiedenheit}.
A large number of sentences can be generated from a finite set of words.
Thus there has been a lot of effort to capture the meaning of words. Some approaches are based on \emph{distributional} word representations~\cite{Lund1996HiDimSemSpacesLex,Patel97extractingsemantic}, others are based on \emph{distributed} representations~\cite{Bengio2008,Collobert2011,Mnih2013,Mikolov2013,Lebret14,pennington2014glove,LevyNIPS2014} where the meaning of a word is encoded as a vector computed from co-occurrence statistics of a word and its neighboring words.
Finally, distributed representations emerged as the solution to many natural language processing (NLP) tasks~\cite{Turney2010,Collobert2011}. 

Given these representations of words in a vector space, techniques for combining them have been proposed to get representations of phrases or sentences.
These compositional models involve vector addition or multiplication~\cite{Mitchell10}. Such simple compositions have shown to perform competitively on the paraphrase detection and phrase similarity tasks~\cite{Blacoe12}.
More sophisticated approaches use techniques from logic, category theory, and quantum information~\cite{Clark2008}. Others use the syntactic relations between words to treat certain words as functions and other as arguments such as adjective-noun composition~\cite{Baroni10} or noun-verb composition~\cite{GrefSadrBarIWCS13}. 
Recursive neural network model for semantic compositionality has also been proposed~\cite{SocherEtAl2012:MVRNN}, where each word has a matrix-vector representation: the vector captures its meaning (as it is initialized with a pre-trained distributed representation), while the matrix learns throught a parse tree how it modifies the meaning of the other word that it combines with.
Many recent works are based on distributed representations of phrases to tackle a wide range of application in NLP: machine translation~\cite{Bahdanau2015}, constituency parsing~\cite{Legrand2015}, sentiment analysis~\cite{Socher2013recursive}, or image captioning~\cite{Lebret15}. 
There is therefore a clear need for distributed word representations that can be easily extrapolated to meaningful phrase representations.

We argue that distributed representation and composition must go hand in hand, i.e., they must be mutually learned.
We present a model that learns to capture meaning of words in distributed representations using a low-rank approximation of a large word co-occurrence matrix.
We choose to stochastically perform this low-rank approximation which enables the model to simultaneously train these representations to compose for producing representations of phrases (see Figure~\ref{fig:joint}).
As composition function, we choose a simple weighted addition for its simplicity and for enabling sequences of words with different lengths to be representated in a common vector space.
Aside from generating distributed representations of words and phrases, this model gives an encoding function (represented by a matrix) which can be used to encode new words or even phrases based on their co-occurrence counts.
This offers two different alternatives for phrase representations:
(1) representation for a query phrase can be inferred by averaging vector representations of its words (only if they all were in the training set), or (2) by using its word co-occurrence statistics. 

Evaluation on the popular word similarity and analogy tasks demonstrate the capability of our joint model for capturing good distributed representations.
We then introduce a novel task for evaluating phrase representations. 
Given a phrase representation, the objective is to retrieve the words that compose the phrase. 
We compare our model against other state-of-the-art methods for distributed word representations which capture meaningful linear substructures~\cite{MikolovICLR2013,pennington2014glove}.
We show that our model achieves similar performance on word evaluation tasks, but that it outperforms other methods on the phrase evaluation task. 

%The paper is organized as follows. Section~\ref{related-works} presents related works. Section~\ref{model} describes the proposed joint model. Section~\ref{exp-results} describes our experimental setup and the results on word and phrase evaluation tasks. Section~\ref{conclusion} concludes.

\section{Related Works}\label{related-works}

In the literature, two major model families exist for learning distributed word representations: the count-based methods and the predictive-based methods.

The count-based methods consist of using the statistical information contained in large corpora of unlabeled text to build large matrices by simply counting words (word co-coocurrence statistics). The rows correspond to words or terms, and the columns correspond to a local context. The context can be documents, such as in latent semantic analysis (LSA)~\cite{Deerwester90}; or other words~\cite{Lund1996HiDimSemSpacesLex}. 
To generate low-dimensional word representations, a low-rank approximation of these large matrices is performed, mainly with a singular value decomposition (SVD). 
Many authors proposed to improve this model with different transformations for the matrix of counts, such as positive pointwise mutual information (PPMI)~\cite{Bullinaria07extractingsemantic,LevyNIPS2014}, or a square root of the co-occurrence probabilities in the form of Hellinger PCA~\cite{Lebret14}.
Instead of using the co-occurrence probabilities, \cite{pennington2014glove} suggest that word vector representations should be learnt with ratios of co-occurrence probabilities.  For this purpose, they introduce a log-bilinear regression model that combines both global matrix factorization and local context window methods.

The predictive-based model has first been introduced as a neural probabilistic language model~\cite{Bengio2003}.
A neural network architecture is trained to predict the next word given a window of preceding words, where words are representated by low-dimensional vector.
Since, some variations of this architecture have been proposed. \cite{Collobert2011} train a language model to discriminate a two-class classification task: if the word in the middle of the input window is related to its context or not. 
More recently, the need of full neural architectures has been questioned~\cite{Mnih2013,MikolovICLR2013}.
\citeauthor{MikolovICLR2013}~(\citeyear{MikolovICLR2013}) propose two predictive-based log-linear models for learning distributed representations of words:
(i) the continous bag-of-words model (CBOW), where the objective is to correctly classify the current (middle) word given a symmetric window of context words around it;
(ii) the skip-gram model, where instead of predicting the current word based on the context, it tries to maximize classification of a word based on another word in the same sentence.
In~\citeauthor{Mikolov2013}~(\citeyear{Mikolov2013}), the authors also introduce a data-driven approach for learning phrases, where the phrases are treated as individual tokens during the training. 

In this paper, we leverage both families: (i) we use the statistical information for learning distributed word representations by approximating the Hellinger PCA with an autoencoder network; (ii) we jointly learn to predict the words that compose a given phrase.

\section{A Joint Model}\label{model}

Some prior works have designed models to learn word representations~\cite{Mnih2013,Mikolov2013,Lebret14}, while others have proposed models to compose these word representations~\cite{Mitchell10,SocherEtAl2012:MVRNN}. We propose instead to jointly learn word representations and their composition by simple summation.

\subsection{Learning Word Representations w.r.t. the Hellinger Distance}

As words occurring in similar contexts tend to have similar meanings~\cite{Harris1954}, word co-occurrence statistics are generally used to embed similar words into a common vector space~\cite{Turney2010}.
Common approaches calculate the frequencies, apply some transformations (tf-idf, PPMI), reduce the dimensionality, and calculate the similarities.
More recently, \citeauthor{Lebret14}~(\citeyear{Lebret14}) proposed a novel method based on a \emph{Hellinger PCA} of the word co-occurrence matrix. They showed that word representations can be learnt even with a reasonable number of context words.
Inspired by this work, we propose to stochastically perform this low-rank approximation. 
For this purpose, we use an autoencoder with only linear activations to find an optimal solution related to the Hellinger PCA~\cite{Bourlard1988}. Replacing the PCA by an autoencoder allows us to learn jointly a cost function which constrains the word information to be kept by summation.

\subsubsection{Word Co-Occurrence Probabilities}
``You shall know a word by the company it keeps"~\cite{Firth57}.
Keeping this famous quote in mind, word co-occurrence probabilities are computed by counting the number of times each context word $c \in \mathcal{D}$ (where $\mathcal{D}\subseteq\mathcal{W}$) occurs around a word $w \in \mathcal{W}$:
\begin{equation}
p(c|w)=\frac{p(c,w)}{p(w)}=\frac{n(c,w)}{\sum_{c_j \in \mathcal{D}}{n(c_j,w)}}\,,
\end{equation}
where $n(c,w)$ is the number of times a context word $c$ occurs in the surrounding of the word $w$.
A multinomial distribution of $|\mathcal{D}|$ classes (words) is thus obtained for each word $w$:
\begin{equation}
P_w=\{p(c_1|w),\ldots,p(c_{|\mathcal{D}|}|w)\}\,.
\end{equation}

\subsubsection{Hellinger Distance}
Similarities between words can be derived by computing a distance between
their corresponding word distributions. Several distances (or metrics) over
discrete distributions exist, such as the Bhattacharyya distance, the
Hellinger distance or Kullback-Leibler divergence. We chose here the
Hellinger distance for its simplicity and symmetry property (as it is a
true distance). Considering two discrete probability distributions $P=(p_1,\ldots,p_k)$
and $Q=(q_1,\ldots,q_k)$, the Hellinger distance is formally defined as:
\begin{equation}
H(P,Q)=\frac{1}{\sqrt{2}}\sqrt{\sum_{i=1}^{k}{(\sqrt{p_i}-\sqrt{q_i})^2}}\,,
\end{equation}
which is directly related to the Euclidean norm of the difference of the square root vectors: 
\begin{equation}
H(P,Q)=\frac{1}{\sqrt{2}}\lVert\sqrt{P}-\sqrt{Q}\rVert_2\,.
\end{equation}
Note that it makes more sense to take the Hellinger distance rather than
the Euclidean distance for comparing discrete distributions, as $P$ and $Q$
are unit vectors according to the Hellinger distance ($\sqrt{P}$ and
$\sqrt{Q}$ are units vector according to the $\ell_2$ norm).

\subsubsection{Autoencoder}
An autoencoder is employed to represent words in a lower dimensional space.
It takes a distribution $\sqrt{P_w}$ as input, encodes it in a more compact representation, and is trained to reconstruct its own input from that representation:
\begin{equation} \label{eq:word}
|| g(f(\sqrt{P_w})) - \sqrt{P_w} ||^2\,,
\end{equation}
where $g(f(\sqrt{P_w}))$ is the output of the network, $f$ is the encoding function which maps distributions in a $m$-dimension (with $m << |\mathcal{D}|$), and $g$ is the decoding function.
$f(\sqrt{P_w})$ is a distributed representation that captures the main factors of variation in the data as the Hellinger PCA does~\cite{Bourlard1988}.
Here, encoder $f \in \mathbb{R}^{m \times |\mathcal{D}|}$ and decoder $g \in \mathbb{R}^{|\mathcal{D}| \times m}$ are both linear layers.
% =  \sqrt{P_w} \times A \times B$, $A \in \mathbb{R}^{|\mathcal{C}| \times m}$ and $A \in \mathbb{R}^{m \times |\mathcal{C}|}$ with $m << |\mathcal{C}|$.

\subsection{Learning to Sum Word Representations}

Interesting compositionality properties have been observed from models based on the addition of representations~\cite{Mikolov2013}. 
An exhaustive comparison of different composition functions has indeed revealed that an additive model performs well on pre-trained word representations~\cite{Mitchell10}.
Because our word representations are learnt from linear operations, the inherent structure of these representations is linear.   
To combine a sequence of words into a common vector space, we then simply apply an element-wise addition of their vector representations.
This approach makes sense and works well when the meaning of a text is literally ``the sum of its parts''. This is usually the case with noun and verb phrase chunks. For example, into phrases such as ``the red cat'' or ``struggle to deal'', each word independently has its proper meaning. 
Distributed representations for such phrase chunks must retain information from the individual words. 
An objective function is thus defined to learn how to combine the word vector representations, while keeping the maximum information from the original vectors.  
An operation as simple as a weighted sum will probably fail for sequences where individual words act as operators that modify the meaning of another word, or for multiword expressions. 
Other more complex functions could be chosen to also include such cases, but we choose to propose a much simpler model (i.e., averaging the word representations) to get phrase chunk representations with unsupervised learning.  
In this paper, we therefore focus on noun and verb phrase chunks.

\subsubsection{Additive Model}
We define $s = (w_1, \ldots, w_T) \in \mathcal{S}$ a phrase chunk of $T$ words, with $\mathcal{S}$ a set of phrase chunks. By feeding all $\sqrt{P_{w}}$ into the autoencoder, a representation $\mathrm{x}_w \in \mathbb{R}^m$ of each word $w \in \mathcal{D}$ is obtained:
\begin{equation}
\mathrm{x}_w = f(\sqrt{P_w})\,.
\end{equation}
By an element-wise addition, a representation of the phrase chunk $s$ can be calculated as:
\begin{equation}
\mathrm{x}_{s} = \frac{1}{T} \sum_{w_t \in s} \mathrm{x}_{w_t}\,.
\end{equation}

%Because $\mathrm{x}_w$ represents the context distribution of words $w$ in a lower dimension, $\mathrm{x}_{s_{1:T}}$ can be seen as the sum of $T$ context distributions.
%All the semantic information coming from the sequence's word distributions must be embedded in its representation.  
%This means that the sequence vector representation $\mathrm{x}_{s_{1:T}}$ must to some extent be similar to its word vector representations $\mathrm{x}_{w_t}$.
%In predictive-based model, such as the Skip-gram model, the objective is to maximize classification of a word based on another word in the same sequence.
%Given the sequence $s_{1:T}$, this training objective can be formulated as:
%\begin{align}
% \frac{1}{T} \sum_{w_t \in s_{1:T}} \sum_{\underset{w_j \neq w_t}{w_j \in s_{1:T}}}  \mathrm{x}_{w_t} \cdot \mathrm{x}_{w_j} &\approx  \sum_{w_j \in s_{1:T}}\Big(\frac{1}{T} \sum_{w_t \in s_{1:T}} \mathrm{x}_{w_t}\Big) \cdot \mathrm{x}_{w_j}\\\nonumber
% &=  \sum_{w_j \in s_{1:T}}\mathrm{x}_{s_{1:T}} \cdot \mathrm{x}_{w_j} 
%\end{align}  

\begin{figure}[!t]
\begin{center}
%\vspace{-.35cm}
%\framebox[4.0in]{$\;$}
\includegraphics[width=\columnwidth]{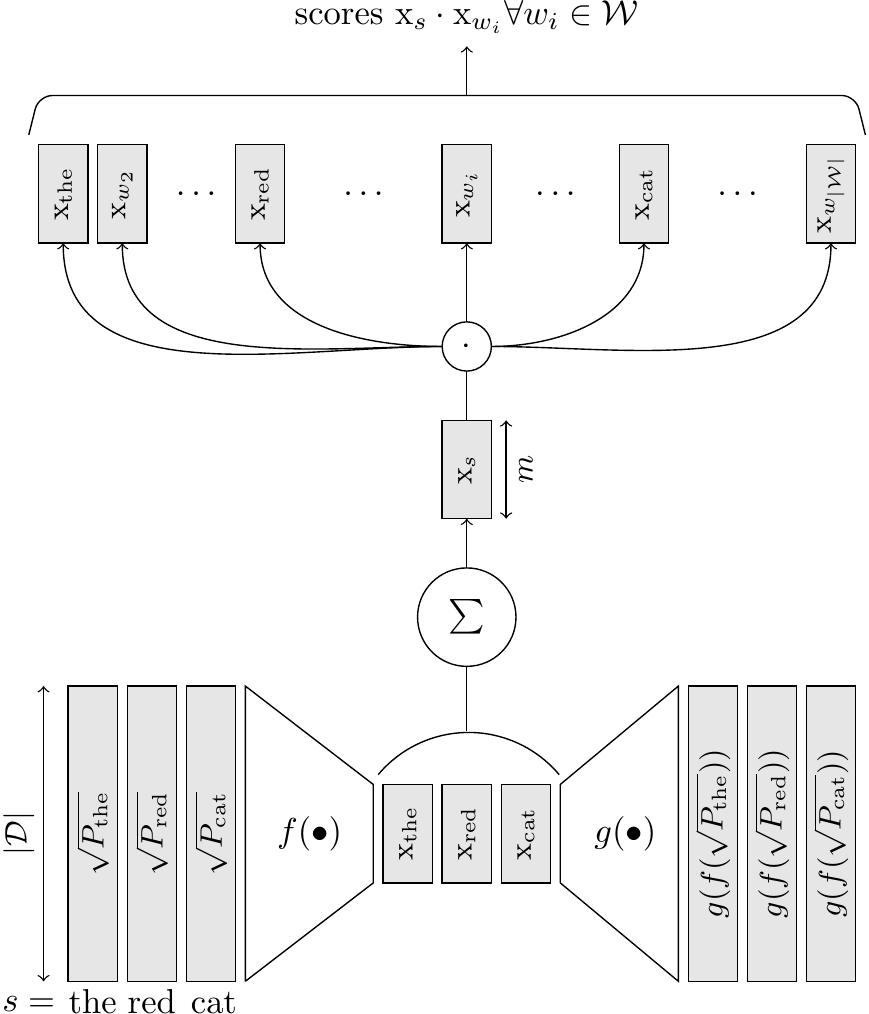}
\end{center}
\caption{Architecture for the joint learning of word representations and their summation. Considering the noun phrase $s=$ \texttt{the red cat}, each word $w_t \in s$ is represented as the square root of its co-occurrence probability distribution $\sqrt{P(w_t)}$. These are the inputs given to an autoencoder which encodes them in a lower dimension $\mathrm{x}_{w_t} \in \mathbb{R}^m$. These new representations are then given to a decoder which is trained to reconstruct the initial inputs. This is the first objective function. The second objective is to keep information when words are summed. All $\mathrm{x}_{w_t}$ are summed together to represent $s$ in the same space as $w_t$. A dot product between the phrase representation $\mathrm{x}_{s}$ and all the other word representations from the dictionary $\mathcal{W}$ is calculated. These scores are trained to be high for words that appear in $s$ and low for the others.}
\label{fig:joint}
\vspace{-.25cm}
\end{figure}

\subsubsection{Training}
In predictive-based model, such as the Skip-gram model, the objective is to maximize the likelihood of a word based on other words in the same sequence.
Instead, our training is slightly different in the sense that we aim at discriminating whether words are in the phrase chunk or not.
An objective function is thus defined to encourage words $w_t$ which appear in the chunk $s$ to give high scores when calculating the dot product between  $\mathrm{x}_{w_t}$ and $\mathrm{x}_{s}$. On the other hand, these scores must be low for words $w_i \notin s$ that do not appear in the chunk.
We train this problem with a ranking-type cost:
\begin{equation} \label{eq:sum}
\sum_{s \in \mathcal{S}} \sum_{w_t \in s} \sum_{\underset{w_i \not \in s}{w_i \in \mathcal{W}}} \operatorname*{max} (0, 1 - \mathrm{x}_{s} \cdot \mathrm{x}_{w_t} + \mathrm{x}_{s} \cdot \mathrm{x}_{w_i})\,.
\end{equation}

Note that due to the large size of $\mathcal{W}$, a negative sampling approach can be used to speed up the training.
In Equation~\ref{eq:sum}, the whole dictionary $\mathcal{W}$ is thus replaced by a subset $\mathcal{W}^{-} \subseteq \mathcal{W}$ with $N$ randomly chosen negative samples $w_i \not \in s$.
A new set $\mathcal{W}^{-}$ is randomly picked at each iteration during the training.

\subsection{Joint Learning}

In contrast with other methods which have subsequently found nice compositionality properties by simple summation, the novelty of our method is the explicit learning of word representations suitable for summation.
The system is then designed to force words with similar context to be close in a $m$-dimensional space, while these dimensions are learnt to be combined with other related words.
This joint learning is illustrated in Figure~\ref{fig:joint}.
The whole system is trained by minimizing both objective functions~\eqref{eq:word} and~\eqref{eq:sum} over the training data using stochastic gradient descent.

\section{Experiments}\label{exp-results}

\subsection{Datasets}

\subsubsection{Building Word Representation over Large Corpora}
\label{wordrep}

Our English corpus is composed of the entire English Wikipedia\footnote{Available at \url{http://download.wikimedia.org}. We took the January 2014 version.} (where all MediaWiki markups have been removed). We consider lower case words to limit the number of words in the dictionary. Additionally, all occurrences of sequences of numbers within a word are replaced with the string ``NUMBER". The resulting text is tokenized using the Stanford tokenizer\footnote{Available at \url{http://nlp.stanford.edu/software/tokenizer.shtml}}. The data set contains about 1.6 billion words.
As dictionary $\mathcal{W}$, we consider all the words within our corpus which appear at least one hundred times. 
This results in a 191,268 words dictionary.
Only the 10,000 most frequent words within this dictionary were used as context words $\mathcal{D}$ to calculate the word co-occurrence probabilities.
A symmetric context window of ten words around each word $w \in \mathcal{W}$ is used to obtain the multinomial distribution $P_w$.
We chose to encode words in a 100-dimensional vector.

\subsubsection{Summing Words for Phrase Representation}
\label{phraserep}
To learn the summation of words that appear frequently together, we choose to consider only the noun and verb phrase chunks to build $\mathcal{S}$. 
We extract these chunks with a phrase chunking approach by using the SENNA software\footnote{Available at \small\url{http://ml.nec-labs.com/senna/}}. 
By retaining only the phrase chunks appearing at least ten times, this results in 1,823,259 noun phrase chunks and 255,232 verb phrase chunks, for a total of 2,078,491 phrase chunks.
We divided this set of phrases into three sets: 1,000 phrases for validation, 5,000 phrases for testing, and the rest for training (2,072,491 phrases).
An unsupervised framework requires a large amount of data. 
Because our primary focus is to provide good word representations, validation and testing sets are intentionally kept small to retain as much phrases as possible in the training set.

\subsection{Other Methods}

We compare our distributed representations with other available models for computing vector representations of words:
(1) the GloVe model which is also based on co-occurrence statistics of corpora~\cite{pennington2014glove}\footnote{Code available at \url{http://www-nlp.stanford.edu/software/glove.tar.gz}.},
(2) the continuous bag-of-words (CBOW) and the skip-gram (SG) architectures which learn representations from prediction-based models~\cite{Mikolov2013}\footnote{Code available at \url{http://word2vec.googlecode.com/svn/trunk/}.}.
The same corpus and dictionary $\mathcal{W}$ as the ones described in Section~\ref{wordrep} are used to train 100-dimensional word vector representations.
We use a symmetric context window of ten words, and the default values set by the authors for the other hyperparameters. 
To see the improvement compared to a standalone SVD, we generate word representations with a truncated SVD of the matrix $X$, where each row of $X$ is a  distribution $\sqrt{P_w}$, $X=\Big( \sqrt{P_{w_1}}, \sqrt{P_{w_2}} , \ldots, \sqrt{P_{w_{|\mathcal{W}|}}} \Big)^{T} \in \mathbb{R}^{|\mathcal{W}| \times |\mathcal{D}|}$.
%\begin{equation}
%X =   \begin{pmatrix}
%  \sqrt{P_{w_1}} \\
%  \sqrt{P_{w_2}} \\
%  \vdots \\
%  \sqrt{P_{w_{|\mathcal{W}|}}}
%  \end{pmatrix}\,.
%\end{equation}

\subsection{Evaluating Word Representations}
\label{wordeval}

The first objective of the model is to learn distributed representations which capture both syntactic and semantic informations about words. 
To evaluate the quality of these representations, we used both analogy and similarity tasks.

\subsubsection{Word analogies}
The word analogy task consists of questions like, ``\emph{a} is to \emph{b} as \emph{c} is to ?''. It was introduced in~\citeauthor{MikolovICLR2013}~(\citeyear{MikolovICLR2013}) and contains 19,544 such questions, divided into a semantic subset and a syntactic subset. The 8,869 semantic questions are analogies about places, like ``\emph{Bern} is to \emph{Switzerland} as \emph{Paris} is to  ?'', or family relationship, like  ``\emph{uncle} is to \emph{aunt} as \emph{boy} is to  ?''. The 10,675 syntactic questions are grammatical analogies, involving plural and adjectives forms, superlatives, verb tenses, etc. To correctly answer the question, the model should uniquely identify the missing term, with only an exact correspondence counted as a correct match.

\subsubsection{Word Similarities}
We also evaluate our model on a variety of word similarity tasks. These include the WordSimilarity-353 Test Collection (WS-353)~\cite{Finkelstein2001}, the Rubenstein and Goodenough dataset (RG-65)~\cite{Rubenstein1965}, and the Stanford Rare Word (RW)~\cite{Luong2013}. They all contain sets of English word pairs along with human-assigned similarity judgements. WS-353 and RG-65 datasets contain 353 and 65 word pairs respectively. Those are relatively common word pairs, like \emph{computer:internet} or \emph{football:tennis}. The RW dataset differs from these two datasets, since it contains 2,034 pairs where one of the word is rare or morphologically complex, such as \emph{brigadier:general} or \emph{cognizance:knowing}.

\subsubsection{Results}

\begin{table}[h]
%\vspace{-.25cm}
\begin{center}
\begin{tabular}{@{}lccccc@{}}\hline\toprule
 & {\bf WS} & {\bf RG}  & {\bf RW} & {\bf SYN.} & {\bf SEM.} \\\bottomrule
\\
CBOW & 0.57 & 0.47 & 0.32 & 53.5 & 22.7\\
Skip-gram &  {\bf0.62} & 0.49 & {\bf0.39} & 66.7 & 53.8\\
GloVe & 0.56 & {\bf0.50} & 0.36 & {\bf79.7} & {\bf75.0} \\\bottomrule
\\
SVD & 0.43 & 0.39 & 0.27 & 52.3 & 34.1  \\
Our model & {\bf0.62} &  0.49 & {\bf0.39} & 69.4 & 43.0 \\\bottomrule
\hline
\end{tabular}
\end{center}
\caption{Word representations evaluation on both similarity and analogy tasks. Comparison of performance across all models with 100-dimensional word vector representations. For all models, a symmetric context window of ten words is used. Spearman rank correlation is reported on word similarity tasks. Accuracy is reported on word analogy tasks.}
\label{wordtable}
%\vspace{-.25cm}
\end{table}

Results reported in Table~\ref{wordtable} show that our model gives similar results than other state-of-the-art methods on word similarity tasks.
However, there is a significant performance boost between the low-rank approximation of $X$ with a SVD and this same approximation with our joint model.
This shows that combining a count-based model with a predictive-based approach helps for generating better word representations.
Performance on word analogy tasks show that our joint model competes with others on the syntactic questions, but that it gives a lower accuracy on semantic questions.
One possible explanation is that less common words are involved in semantic questions compared to syntactic questions. Among the four words that make a semantic question, one of them is, in average, the $34328^{th}$ most frequent word in $\mathcal{W}$, while it is the $20819^{th}$ for a syntactic question.    
Compared to other methods which take the whole dictionary $\mathcal{W}$ as context dictionary, we consider only a small subset of it ($\mathcal{D}$ contains only the $10000$ most frequent words of $\mathcal{W}$). A larger context dictionary would certainly help to improve performance on this task\footnote{It has not been explored due to limitations in hardware resources. It would be easily computable with a cluster of CPU.}.

\subsection{Evaluating Phrase Representations}
\label{phraseval}

As a second objective, we aim at learning to sum word representations to generate phrase representations while keeping the original information coming from the words. 
We thus introduce a novel task to evaluate the phrase representations. 

\subsubsection{Description of the Task}
As dataset, we use the collection of test phrases described in Section~\ref{phraserep}.
It contains 5000 phrases (noun phrases and verb phrases) extracted from Wikipedia with a chunking approach. Amoung them, 2244, 2030 and 547 are, respectively, composed of two, three and four words. The remaining 179 are composed of at least five words with a maximum of eight words.
For a given phrase $s = (w_1, \ldots, w_T) \in \mathcal{S}$ of $T$ words, the objective is to retrieve the $T$ words from its distributed representation $\mathrm{x}_{s}$. 
Scores between the phrase $s$ and all the possible words $w_i \in \mathcal{W}$ are calculated using the dot product between their distributed representations $\mathrm{x}_{s} \cdot \mathrm{x}_{w_i}$, as illustrated in Figure~\ref{fig:joint}.
The top $T$ scores are considered as the words composing the phrase $s$.

\subsubsection{Results}

\begin{figure}[!t]
%\vspace{-0.5cm}
  \includegraphics[width=\columnwidth]{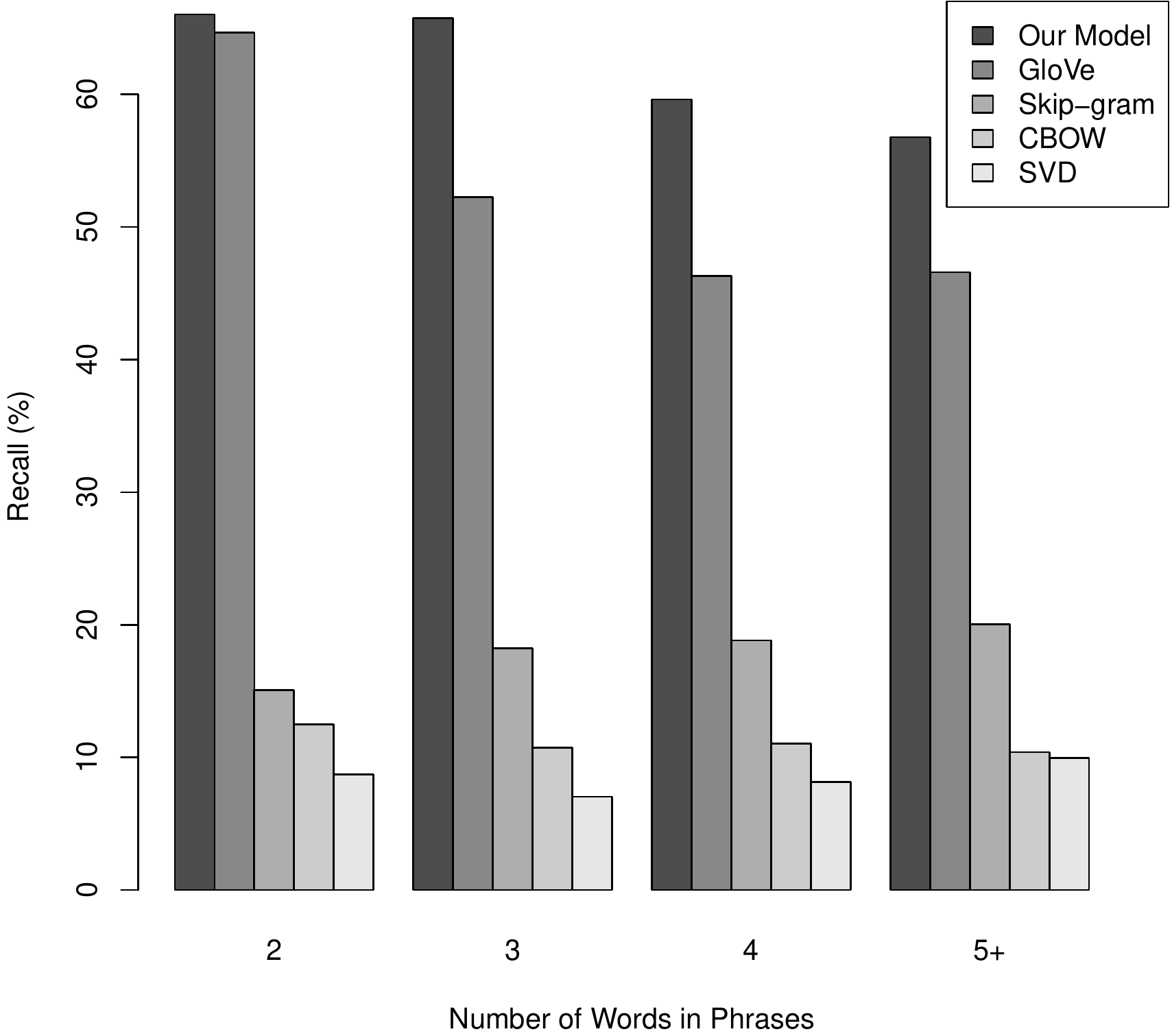}
  \caption{Recall@1 based on the number of words per phrases. Comparison of performance across all models with 100-dimensional word vector representations.}
  \label{fig:stats}
%\vspace{-0.3cm}
\end{figure}

To evaluate whether words making a given phrase can be retrieved from the distributed phrase representation, we use Recall @$K$, which measures the fraction of times a correct word was found among the top $K$ results. $K$ is proportional to the number of words per phrase, e.g.\ for a 3-word phrase with a Recall@5, the correct words are found amoung the top 15 results.
Higher Recall @$K$ means better retrieval performance. Since we care most about the top-ranked retrieved results, the Recall @$K$ with small $K$ are more important. 

\begin{table}[!h]
%\vspace{-.25cm}
\begin{center}
\begin{tabular}{@{}lccc@{}}\hline\toprule
 & {\bf R@1} & {\bf R@5}  & {\bf R@10} \\\bottomrule
\\
CBOW & 11.33 & 29.56 & 38.46 \\
Skip-gram & 7.96 & 22.26 & 30.04 \\
GloVe & 54.97 & 79.97 & 86.54 \\\bottomrule
\\
SVD & 17.42 & 32.87 & 40.72 \\
Our model & {\bf64.22} & {\bf91.72} & {\bf95.85} \\\bottomrule
\hline
\end{tabular}
\end{center}
\caption{Phrase representations evaluation. Comparison of performance across all models with 100-dimensional phrase vector representations on word retrieval. R@$K$ is Recall@$K$, with $K=\{1,5,10\}$.}
\label{phrasetable}
%\vspace{-.25cm}
\end{table}

\begin{table*}[!t]
%\vspace{-0.2cm}
\begin{center}
\resizebox{0.9\textwidth}{!}{
\begin{tabular}{@{}llll@{}}\hline\toprule
{\bf QUERY PHRASES} & \multicolumn{3}{c}{\bf NEAREST PHRASES} \\
\cmidrule{2-4}	
& \multicolumn{1}{c}{\sc Encoding Function $f$} & \phantom{a} &  \multicolumn{1}{c}{\sc Averaging Words} \\\midrule
\\
 \multirow{5}{6cm}{\bf\sc american airlines} & \sc\small braniff airlines & & \sc\small  american airways \\
 & \sc\small aloha airlines & &  \sc\small pan american airlines\\
 & \sc\small braniff airways & &  \sc\small american eagle airlines \\
 & \sc\small jetblue airways & &  \sc\small north american airlines \\
 & \sc\small braniff international airways & &  \sc\small american overseas airlines\\\bottomrule
 \\
 \multirow{5}{6cm}{\bf\sc chicago bulls} & \sc\small denver nuggets & &  \sc\small chicago colts \\
 & \sc\small seattle supersonics & &  \sc\small chicago hornets\\
 & \sc\small cleveland cavaliers & &  \sc\small chicago stags\\
 & \sc\small boston celtics  & &  \sc\small buffalo bulls \\
 & \sc\small dallas mavericks & &  \sc\small chicago cardinals \\\bottomrule
 \\
 \multirow{5}{6cm}{\bf\sc home plate} & \sc\small  right fielder  & &  \sc\small the home plate umpire \\
 & \sc\small center fielders & &  \sc\small the home plate area\\
 & \sc\small the outfield fence & &  \sc\small the home leg \\
 & \sc\small leadoff batter & &  \sc\small the ball home \\
 & \sc\small the infield & &  \sc\small the diamond state base ball club \\\bottomrule
 \\
 \multirow{5}{6cm}{\bf\sc president of the united states} & \sc\small president coolidge & &  \sc\small  the united states president\\
 & \sc\small president eisenhower & &  \sc\small the united states presidency \\
 & \sc\small u.s. president dwight eisenhower & &  \sc\small the first united states secretary\\
 & \sc\small president truman & &  \sc\small the united states minister \\
 & \sc\small president reagan & &  \sc\small the first united states senator \\\bottomrule
\end{tabular}}
\end{center}
\caption{Examples of phrases and five of their ten nearest phrases from the collection of phrases. Representations for the collection of phrases have been computed by averaging the word representations. Query phrase representations are inferred using the two different alternatives: (1) with the encoding function $f$ using counts from a symmetric window of ten context words around the query phrase, (2) by averaging the representations of the words that compose the query phrase. All distributed representations are 100-dimensional vectors.}
\label{neigh-table2}

%\vspace{-0.1cm}
\end{table*}
Results reported in Table~\ref{phrasetable} show that our distributed word representations can be averaged together to produce meaningful phrase representations, since the words are retrieved with a high recall.
Our model significantly outperforms others methods on this task. 
In Figure~\ref{fig:stats}, a more detailed analysis of results reveals that the GloVe model competes with ours for the 2-word phrases.
However GloVe's representations cannot maintain this performance for longer phrases. It is probably not too surprising as this model is trained using ratios of co-occurrence probabilities for two target words.
Consequently, it well learns linear substructures for pairs of words, which probably also explains its good performance on word analogy tasks. 
In contrast, our joint model can learn more complex substructures which make possible the aggregation of multiple words within a low-dimensional vector space.

\subsection{Inferring New Phrase Representations}

%We train a model able to combine word representations to produce phrase representations. 
Representations for new phrases can thus be generated by simply averaging its word representations, assuming that the words are in the dictionary $\mathcal{W}$.
Consider that the dictionary $\mathcal{W}^n$ tends to grow exponentially with $n$, it gives a nice framework to produce the huge variety of possible sequence of $n$ words in a timely and efficient manner with low memory consumption, unlike other methods.
Relying on word co-occurrence statistics to represent words in vector space also provides a framework to easily generate representations for unseen words.
This is another advantage compared to methods focused on learning distributed word representations (such as CBOW, Skip-gram and GloVe models), where the whole system needs to be trained again to learn representations for these new constituents.
To infer a representation for a new word $w_{\text{new}}$, one only needs to count its context words over a large corpus of text to build the distribution $\sqrt{P_{w_{\text{new}}}}$.
This nice feature can be extrapolated to phrases, which gives another alternative for generating phrase representations. 
Table~\ref{neigh-table2} presents some examples of phrases, where we use both alternatives to compute their distributed representations.
%There are some interesting query phrases where the meaning clearly depends on the composition of their words. For instance, words from the entity \emph{Chicago Bulls} differ in meaning when taken separately. \emph{Chicago} should be close to other american cities, and \emph{Bulls} should be close to other horned animals.
It can be seen that both alternatives give distinct representations.
For instance, by using the encoding function $f$, our model infers a representation for the entity \emph{Chicago Bulls} which is close to other NBA teams, like the \emph{Denver Nuggets} or the \emph{Seattle Supersonics}. 
By averaging the representations of both words \emph{Chicago} and \emph{Bulls}, our model infers a representation which is close to other Chicago's sport teams.
Both representations are meaningful, but they carry different information.
Relying on co-occurrence statistics gives entities that occur in a similar context, while the summation tries to find entities containing the maximum amount of similar information.  
This also works with longer phrases, such as \emph{President of the United States}.
The first alternative gives men who served as president, when the second gives related positions. 

\section{Conclusion}\label{conclusion}

We introduce a model that combines both count-based methods and predictive-based methods for generating distributed representations of words and phrases.
Using a chunking approach, a collection of noun phrases and verb phrases is extracted from Wikipedia.
For a given $n$-word phrase, we train our model to generate a low-dimensional representation for each word based on its co-occurrence probability distribution.
These $n$ representations are averaged together to generate a distributed phrase representation in the same semantic space.
Thanks to an autoencoder approach, we can simultaneously train the model to retrieve the original $n$ words from the phrase representation, and therefore learn complex linear substructures.
When compared to state-of-the-art methods on some classical word evaluation tasks, the competitive results show that our joint model produces meaningful word representations.
Performance on a novel task for evaluating phrase representations confirm the ability of our model to learn complex substructures, which make possible the aggregation of multiple words within a low-dimensional vector space.  
Better still, inference of new phrase representations is also easily feasible when relying on counts. Some quantitative examples demonstrate that both alternatives can give different but meaningful information about phrases.  
The word representations and the collection of phrases used in these experiments are available online, here: \url{http://www.lebret.ch/words/}.

\section*{Acknowledgements}

This work was supported by the HASLER foundation through the grant ``Information and Communication Technology for a Better World 2020'' (SmartWorld).

%%%%%%%%%%%%%%%%%%%%%%%%%%%%%%%%%
%% Bibliography
%%%%%%%%%%%%%%%%%%%%%%%%%%%%%%%%%
\bibliography{icml2015}

\begin{thebibliography}{31}
\providecommand{\natexlab}[1]{#1}
\providecommand{\url}[1]{\texttt{#1}}
\expandafter\ifx\csname urlstyle\endcsname\relax
  \providecommand{\doi}[1]{doi: #1}\else
  \providecommand{\doi}{doi: \begingroup \urlstyle{rm}\Url}\fi

\bibitem[Bahdanau et~al.(2015)Bahdanau, Cho, and Bengio]{Bahdanau2015}
Bahdanau, Dzmitry, Cho, Kyunghyun, and Bengio, Yoshua.
\newblock Neural machine translation by jointly learning to align and
  translate.
\newblock In \emph{Proceedings of the 3th International Conference on Learning
  Representations (ICLR)}, 2015.

\bibitem[Baroni \& Zamparelli(2010)Baroni and Zamparelli]{Baroni10}
Baroni, M. and Zamparelli, R.
\newblock {Nouns are Vectors, Adjectives are Matrices: Representing
  Adjective-Noun Constructions in Semantic Space}.
\newblock In \emph{Proceedings of the EMNLP}, pp.\  1183--1193, 2010.

\bibitem[Bengio(2008)]{Bengio2008}
Bengio, Y.
\newblock {Neural net language models}.
\newblock \emph{Scholarpedia}, 3\penalty0 (1):\penalty0 3881, 2008.

\bibitem[Bengio et~al.(2003)Bengio, Ducharme, Vincent, and Janvin]{Bengio2003}
Bengio, Y., Ducharme, R., Vincent, P., and Janvin, C.
\newblock {A neural probabilistic language model}.
\newblock \emph{The Journal of Machine Learning Research}, 3:\penalty0
  1137--1155, 2003.

\bibitem[Blacoe \& Lapata(2012)Blacoe and Lapata]{Blacoe12}
Blacoe, W. and Lapata, M.
\newblock {A Comparison of Vector-based Representations for Semantic
  Composition}.
\newblock In \emph{Proceedings of the 2012 Joint Conference on Empirical
  Methods in Natural Language Processing and Computational Natural Language
  Learning}, EMNLP-CoNLL '12, pp.\  546--556. Association for Computational
  Linguistics, 2012.

\bibitem[Bourlard \& Kamp(1988)Bourlard and Kamp]{Bourlard1988}
Bourlard, H. and Kamp, Y.
\newblock Auto-association by multilayer perceptrons and singular value
  decomposition.
\newblock \emph{Biological cybernetics}, 59\penalty0 (4-5):\penalty0 291--294,
  1988.

\bibitem[Bullinaria \& Levy(2007)Bullinaria and
  Levy]{Bullinaria07extractingsemantic}
Bullinaria, J.~A. and Levy, J.~P.
\newblock {Extracting semantic representations from word co-occurrence
  statistics: A computational study}.
\newblock \emph{Behavior Research Methods}, 39\penalty0 (3):\penalty0 510--526,
  2007.

\bibitem[Clark et~al.(2008)Clark, Coecke, and Sadrzadeh]{Clark2008}
Clark, S., Coecke, B., and Sadrzadeh, M.
\newblock A compositional distributional model of meaning.
\newblock In \emph{Proceedings of the Second Quantum Interaction Symposium
  (QI-2008)}, pp.\  133--140, 2008.

\bibitem[Collobert et~al.(2011)Collobert, Weston, Bottou, Karlen, Kavukcuoglu,
  and Kuksa]{Collobert2011}
Collobert, R., Weston, J., Bottou, L., Karlen, M., Kavukcuoglu, K., and Kuksa,
  P.
\newblock {Natural Language Processing (Almost) from Scratch}.
\newblock \emph{The Journal of Machine Learning Research}, 12:\penalty0
  2493--2537, 2011.

\bibitem[Deerwester et~al.(1990)Deerwester, Dumais, Furnas, Landauer, and
  Harshman]{Deerwester90}
Deerwester, S., Dumais, S.~T., Furnas, G.~W., Landauer, T.~K., and Harshman, R.
\newblock Indexing by latent semantic analysis.
\newblock \emph{Journal of the American Society for Information Science},
  41\penalty0 (6):\penalty0 391--407, 1990.

\bibitem[Finkelstein et~al.(2001)Finkelstein, Gabrilovich, Matias, Rivlin,
  Solan, Wolfman, and Ruppin]{Finkelstein2001}
Finkelstein, L., Gabrilovich, E., Matias, Y., Rivlin, E., Solan, Z., Wolfman,
  G., and Ruppin, E.
\newblock {Placing Search in Context: The Concept Revisited}.
\newblock In \emph{Proceedings of the 10th international conference on World
  Wide Web}, pp.\  406--414. ACM, 2001.

\bibitem[Firth(1957)]{Firth57}
Firth, J.~R.
\newblock {A Synopsis of Linguistic Theory 1930-55.}
\newblock 1957.

\bibitem[Grefenstette et~al.(2013)Grefenstette, Dinu, Zhang, Sadrzadeh, and
  Baroni]{GrefSadrBarIWCS13}
Grefenstette, E., Dinu, G., Zhang, Y., Sadrzadeh, M., and Baroni, M.
\newblock \emph{Proceedings of the 10th International Conference on
  Computational Semantics (IWCS 2013) -- Long Papers}, chapter Multi-Step
  Regression Learning for Compositional Distributional Semantics, pp.\
  131--142.
\newblock Association for Computational Linguistics, 2013.

\bibitem[Harris(1954)]{Harris1954}
Harris, Z.~S.
\newblock Distributional structure.
\newblock \emph{Word}, 1954.

\bibitem[Humboldt(1836)]{humboldt1836verschiedenheit}
Humboldt, W.
\newblock \emph{{\"Uber die Verschiedenheit des menschlichen Sprachbaues: Und
  ihren Einfluss auf die geistige Entwickelung des Menschengeschlechts}}.
\newblock Druckerei der K{\"o}niglichen Akademie der Wissenschaften, 1836.

\bibitem[Lebret \& Collobert(2014)Lebret and Collobert]{Lebret14}
Lebret, R. and Collobert, R.
\newblock {Word Embeddings through Hellinger PCA}.
\newblock In \emph{Proceedings of the 14th Conference of the European Chapter
  of the Association for Computational Linguistics}, pp.\  482--490.
  Association for Computational Linguistics, April 2014.

\bibitem[Lebret et~al.(2015)Lebret, Pinheiro, and Collobert]{Lebret15}
Lebret, R., Pinheiro, P. H.~O., and Collobert, R.
\newblock Phrase-based image captioning.
\newblock In \emph{Proceedings of the 32st International Conference on Machine
  Learning (ICML)}, 2015.

\bibitem[Legrand \& Collobert(2015)Legrand and Collobert]{Legrand2015}
Legrand, J. and Collobert, R.
\newblock Joint rnn-based greedy parsing and word composition.
\newblock In \emph{Proceedings of the 3th International Conference on Learning
  Representations (ICLR)}, 2015.

\bibitem[Levy \& Goldberg(2014)Levy and Goldberg]{LevyNIPS2014}
Levy, O. and Goldberg, Y.
\newblock {Neural Word Embedding as Implicit Matrix Factorization}.
\newblock In \emph{Advances in Neural Information Processing Systems}, pp.\
  2177--2185. 2014.

\bibitem[Lund \& Burgess(1996)Lund and Burgess]{Lund1996HiDimSemSpacesLex}
Lund, K. and Burgess, C.
\newblock {Producing high-dimensional semantic spaces from lexical
  co-occurrence}.
\newblock \emph{Behavior Research Methods, Instruments, \& Computers},
  28\penalty0 (2):\penalty0 203--208, 1996.

\bibitem[Luong et~al.(2013)Luong, Socher, and Manning]{Luong2013}
Luong, M., Socher, R., and Manning, C.~D.
\newblock {Better Word Representations with Recursive Neural Networks for
  Morphology}.
\newblock In \emph{CoNLL}, 2013.

\bibitem[Mikolov et~al.(2013{\natexlab{a}})Mikolov, Chen, Corrado, and
  Dean]{MikolovICLR2013}
Mikolov, T., Chen, K., Corrado, G., and Dean, J.
\newblock {Efficient Estimation of Word Representations in Vector Space}.
\newblock \emph{In Proceedings of Workshop at International Conference on
  Learning Representations (ICLR 2013)}, 2013{\natexlab{a}}.

\bibitem[Mikolov et~al.(2013{\natexlab{b}})Mikolov, Sutskever, Chen, Corrado,
  and Dean]{Mikolov2013}
Mikolov, T., Sutskever, I., Chen, K., Corrado, G., and Dean, J.
\newblock {Distributed Representations of Words and Phrases and their
  Compositionality}.
\newblock In \emph{Advances in Neural Information Processing Systems}, pp.\
  3111--3119. 2013{\natexlab{b}}.

\bibitem[Mitchell \& Lapata(2010)Mitchell and Lapata]{Mitchell10}
Mitchell, J. and Lapata, M.
\newblock {Composition in Distributional Models of Semantics}.
\newblock \emph{Cognitive Science}, 34\penalty0 (8):\penalty0 1388--1429, 2010.

\bibitem[Mnih \& Kavukcuoglu(2013)Mnih and Kavukcuoglu]{Mnih2013}
Mnih, A. and Kavukcuoglu, K.
\newblock {Learning word embeddings efficiently with noise-contrastive
  estimation}.
\newblock In \emph{Advances in Neural Information Processing Systems}, pp.\
  2265--2273, 2013.

\bibitem[Patel et~al.(1998)Patel, Bullinaria, and
  Levy]{Patel97extractingsemantic}
Patel, M., Bullinaria, J.~A., and Levy, J.~P.
\newblock {Extracting Semantic Representations from Large Text Corpora}.
\newblock In \emph{4th Neural Computation and Psychology Workshop, London,
  9--11 April 1997}, pp.\  199--212. Springer, 1998.

\bibitem[Pennington et~al.(2014)Pennington, Socher, and
  Manning]{pennington2014glove}
Pennington, J., Socher, R., and Manning, C.~D.
\newblock {GloVe: Global Vectors for Word Representation}.
\newblock In \emph{Proceedings of the Empiricial Methods in Natural Language
  Processing (EMNLP 2014)}, volume~12, 2014.

\bibitem[Rubenstein \& Goodenough(1965)Rubenstein and
  Goodenough]{Rubenstein1965}
Rubenstein, H. and Goodenough, J.~B.
\newblock {Contextual Correlates of Synonymy}.
\newblock \emph{Communications of the ACM}, 8\penalty0 (10):\penalty0 627--633,
  1965.

\bibitem[Socher et~al.(2012)Socher, Huval, Manning, and
  Ng]{SocherEtAl2012:MVRNN}
Socher, R., Huval, B., Manning, C., and Ng, A.
\newblock {Semantic Compositionality Through Recursive Matrix-Vector Spaces}.
\newblock In \emph{Proceedings of the 2012 Joint Conference on Empirical
  Methods in Natural Language Processing and Computational Natural Language
  Learning}, pp.\  1201--1211. Association for Computational Linguistics, 2012.

\bibitem[Socher et~al.(2013)Socher, Perelygin, Wu, Chuang, Manning, Ng, and
  Potts]{Socher2013recursive}
Socher, R., Perelygin, A., Wu, J.~Y., Chuang, J., Manning, C.~D., Ng, A.~Y, and
  Potts, C.
\newblock {Recursive Deep Models for Semantic Compositionality Over a Sentiment
  Treebank}.
\newblock In \emph{Proceedings of the 2013 Conference on Empirical Methods in
  Natural Language Processing (EMNLP)}, volume 1631, pp.\  1642. Association
  for Computational Linguistics, 2013.

\bibitem[Turney \& Pantel(2010)Turney and Pantel]{Turney2010}
Turney, P. and Pantel, P.
\newblock {From Frequency to Meaning: Vector Space Models of Semantics}.
\newblock \emph{Journal of Artificial Intelligence Research}, 37\penalty0
  (1):\penalty0 141--188, 2010.

\end{thebibliography}
\bibliographystyle{icml2015}

\end{document}